\documentclass[10pt,twocolumn,letterpaper]{article}

\usepackage{iccv}
\usepackage{times}
\usepackage{epsfig}
\usepackage{graphicx}
\usepackage{amsmath}
\usepackage{amssymb}

\usepackage{epsfig}
\usepackage{graphicx}
\usepackage{amsmath}
\usepackage{amssymb}
\usepackage{booktabs}
\usepackage{xcolor}
\usepackage{algorithm,algorithmic}

\newcommand{\V}[1]{{\mathbf{#1}}}
\newcommand{\M}[1]{{\ensuremath\mathbf{#1}}}
\newcommand{\pnp}{P\&P}

\usepackage[pagebackref=true,breaklinks=true,letterpaper=true,colorlinks,bookmarks=false]{hyperref}

\iccvfinalcopy %

\ificcvfinal\fi

\begin{document}

\title{Just Label What You Need: Fine-Grained Active Selection for \\
Perception and Prediction through Partially Labeled Scenes}

\author{
Sean Segal$^{12}$\thanks{Equal Contribution}\hspace{0.1cm}, Nishanth Kumar$^{3}$\footnotemark[1]\hspace{0.15cm}\thanks{Work done while at Uber ATG}, Sergio Casas$^{12}$, Wenyuan Zeng$^{12}$ \\
Mengye Ren$^{12}$, Jingkang Wang$^{12}$, Raquel Urtasun$^{12}$\\
University of Toronto$^1$, Uber ATG$^2$, Brown University$^{3}$\\
{\tt\small \{seansegal, sergio, wenyuan, mren, wangjk, urtasun\}@cs.toronto.edu, nishanth\_kumar@brown.edu}
}

\maketitle
\ificcvfinal\thispagestyle{empty}\fi

\begin{abstract}
\vspace{-0.3cm}
Self-driving vehicles must perceive and predict the future positions of nearby actors in order to avoid collisions and drive safely. 
A learned deep learning module is often responsible for this task, requiring large-scale, high-quality training datasets. As data collection is often significantly cheaper than labeling in this domain, the decision of which subset of examples to label can have a profound impact on model performance. 
Active learning techniques, which leverage the state of the current model to iteratively select examples for labeling, offer a promising solution to this problem.
However, despite the appeal of this approach, there has been little scientific analysis of active learning approaches for the perception and prediction (\pnp{}) problem. 
In this work, we study active learning techniques for \pnp{} and find that the traditional active learning formulation is ill-suited for the \pnp{} setting.  
We thus introduce generalizations that ensure that our approach is both cost-aware and allows for fine-grained selection of examples through partially labeled scenes. 
Our experiments on a real-world, large-scale self-driving dataset suggest that fine-grained selection can improve the performance across perception, prediction, and downstream planning tasks. 
\end{abstract}

\vspace{-0.3cm}
\section{Introduction}

In order for self-driving vehicles to safely plan a route, 
they must perceive nearby actors and predict their future locations.
In a self-driving stack, a learned perception and prediction (\pnp{}) model is responsible for this task, taking raw sensor data as input and producing object detections and future predictions.
These models typically require large-scale, high-quality training datasets for best performance due to the high dimensional sensor inputs and long tail of possible outcomes that must be learned.
While self-driving companies collect massive amounts of data from real-world driving, annotating these scenes remains a major bottleneck.
Furthermore, some of the collected data may be less interesting for model training -- e.g., a prediction dataset with many parked vehicles is less informative than one with many highly interactive, moving actors. 
As a consequence, the choice of which examples to label is crucial to maximize performance for a given budget.

\begin{figure}[t]
    \centering
    \vspace{-0.3cm}
    \includegraphics[width=1.0\linewidth]{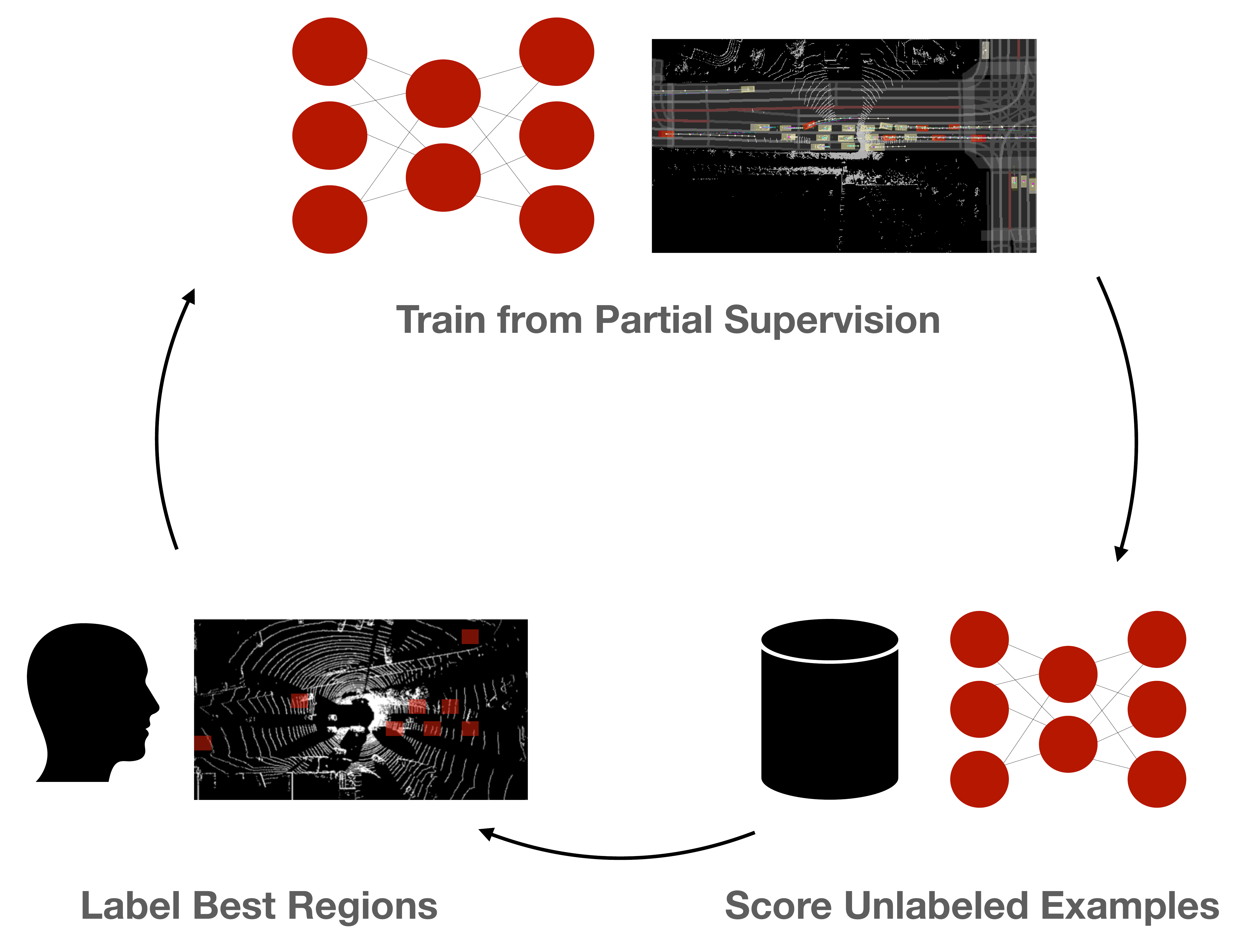}
    \caption{{\bf Fine-Grained Active Selection:} Train on labeled regions (top), score remaining unlabeled examples (bottom right), query labels for the highest scoring regions (bottom left). }
    \label{fig:teaser_figure}
    \vspace{-0.3cm}
\end{figure}

Given a particular model, it is natural to ask whether
it can be used to determine examples most likely to improve performance when labeled.
This problem is well-studied in the field of active learning and recent work has shown impressive performance gains over random selection in many tasks, including image classification, semantic segmentation and 2D object detection \cite{Yoo2019LearningLF, sinha_variational_2019, Agarwal2020ContextualDF}.
Active learning presents a promising framework to employ in real-world self-driving development, where models can continually improve as new batches of examples are selected iteratively for labeling (see Figure~\ref{fig:teaser_figure}).

Despite the appeal of active learning, few approaches have been developed for the self-driving domain.
Scientific analysis is limited to object detection \cite{Haussmann2020ScalableAL, feng2019deep}, with no approaches designed for \pnp{}.
When applying active selection to this task, we find the traditional active learning formulation to be ill-suited.
First, while approaches typically assume fixed labeling costs per example, annotation costs can vary drastically as the cost is highly dependent on the number of actors present. 
Furthermore, the spatial label structure can be exploited to support partial labeling, allowing for fine-grained active selection. Specifically, with small modifications, \pnp{} models can be trained from partial supervision. This enables the active learner to select specific actors in a scene without requiring the remaining actors to be selected, as they may be uninteresting for model improvement. 
With these differences in mind, we introduce a fine-grained, cost-aware selection  along with specific scoring criteria for active selection in the \pnp{} setting.

We leverage a real-world, large-scale dataset to analyze the effects of partial labeling and active selection for \pnp{}.
First, we study models trained on partial supervision, without active selection, and observe that sparse labeling leads to better performance, while trading off model training time, as models require more training iterations when supervision is less dense.
Next, we analyze the additional benefits of fine-grained active selection for prediction and perception. 
Our results demonstrate that fine-grained selection with a simple prediction entropy selection criterion outperforms a variety of popular active learning approaches. 
We further analyze prediction performance broken down by  action, demonstrating the gains are most significant on rare events. 
Ultimately, %
we evaluate the effects on downstream motion planning and show significant improvements for fine-grained active selection.
We also see similar improvements when optimizing perception performance using detection entropy for selection.
All together, our analysis suggests that the dominant paradigm of labeling entire self-driving scenes is not the most efficient use of a fixed labeling budget and that more fine-grained active selection may be required to most effectively select examples for labeling.
\section{Related Work}

\paragraph{Perception and Prediction:}
While perception and prediction have traditionally been handled separately, \cite{Luo2018FastAF, Casas2018IntentNetLT} introduce models to jointly perform both tasks, improving performance and efficiency. 
Among the exciting progress made in both tasks over recent years, most relevant to our work are improvements in prediction representations, allowing models to better characterize uncertainty.  
Examples of representations include trajectories \cite{Casas2018IntentNetLT, PhanMinh2020CoverNetMB}, probabilistic occupancy maps \cite{jain2020discrete, oh2020hcnaf}, Gaussian mixtures \cite{cui2019multimodal}, implicit latent variable models \cite{casas2020implicit}, and auto-regressive models \cite{Tang2019MultipleFP}.

\paragraph{Active Learning:} We focus on pool-based active learning, in which new training examples are queried from a large, unlabeled pool \cite{settles2009active}. 
One class of approaches seeks to characterize model uncertainty, measured via model disagreement \cite{lakshminarayanan2016simple, Beluch2018ThePO}, entropy \cite{Gal2017DeepBA, holub2008entropy}, a learned loss prediction \cite{Yoo2019LearningLF}, 
or a discriminator score \cite{Sinha2019VariationalAA},
and select examples with high uncertainty for labeling. 
While often effective, uncertainty based approaches can also be prone to selecting a subset of similar examples when computational constraints require large batches of examples to be selected before retraining.
This motivates diversity-based approaches \cite{nguyen2004active, Sener2018ActiveLF, guo2010active, gudovskiy2020deep} which seek to find a diverse subset of the unlabeled pool that best characterizes the entire pool. 
\cite{Agarwal2020ContextualDF, Ash2020DeepBA} introduce approaches which balance both uncertainty and diversity in the selection process. 
Most related to our domain are \cite{Haussmann2020ScalableAL, feng2019deep, sivaraman2014active, roy2018deep} which study active learning approaches specifically for object detection.
While most approaches assume fixed labeling costs per example, \cite{Wang2019, wang_cost-effective_2017} have explored explicitly modeling individual labeling costs as part of the selection process. 
Leveraging partially labeled data for fine-grained active selection has been explored in semantic segmentation \cite{Kasarla2019RegionbasedAL, Mackowiak2018CEREALSC} and more generally, in the context of structured prediction problems \cite{Luo2013LatentSA}.

\paragraph{Dataset Selection:} 
In practice, many self-driving datasets select examples manually \cite{caesar2020nuscenes}, randomly, or via hardcoded rules.
Argoverse describes rules-based criteria to mine interesting trajectories for prediction \cite{chang2019argoverse}. 
More recently, \cite{sadat2021diverse} proposed a set of complexity measures for dataset selection. 
Finally, \cite{segal2020universal} proposed tagging attributes of self-driving scenes, enabling log retrieval for dataset curation.

\section{Active Learning for \pnp{}}
Given the high costs associated with labeling perception and prediction datasets, there is a significant opportunity to spend budgets more efficiently by selecting only the best examples for labeling. 
Active learning offers a promising solution, selecting examples predicted most likely to improve model performance.
In this section, we first review the traditional pool-based active learning formulation, where examples are iteratively selected from an  unlabeled pool to incrementally build a high-quality labeled dataset.
Then, we address this formulation's shortcomings in the \pnp{} setting by introducing a new paradigm which is both cost-aware and enables fine-grained selection though partially labeled scenes, 
providing flexibility to ensure labeling budgets are spent most effectively.
Finally, we provide concrete selection criteria which can be used within our framework to optimize the model's perception and prediction performance. 

\begin{algorithm}[t]
    \caption{Active Learning Selection}
    \label{algorithm:traditional_selection}
    \hspace*{\algorithmicindent} \textbf{Input:} \\
    \hspace*{\algorithmicindent} \hspace*{\algorithmicindent} Unlabeled pool: $X_{U}$ , Initial labels: $X_{L}^{(0)}$ \\
    \hspace*{\algorithmicindent} \hspace*{\algorithmicindent} Budget: $K$,  Iterations: $N$, Model: $\mathcal{M}$ \\
    \hspace*{\algorithmicindent} \textbf{Output:} \\
    \hspace*{\algorithmicindent} \hspace*{\algorithmicindent} 
    Final dataset: $X^{(N)}_{L}$, Optimized Model: $\mathcal{M}$ \\
    \begin{algorithmic}[1]
        \FOR{$i \in 1 \dots N$}
            \STATE $\mathcal{S}^{(i)} \leftarrow $ \texttt{score($\mathcal{M}, X_U \setminus X_{L}^{(i - 1)}$)}
            \STATE $Q^{(i)} \leftarrow \texttt{select\_top\_k}(\mathcal{S}^{(i)}, K)$
            \STATE $X_{L}^{(i + 1)} \leftarrow X_{L}^{(i)} \cup Q^{(i)}$
            \STATE $\mathcal{M} \leftarrow \texttt{train}(\mathcal{M}, X_{L}^{(i + 1)})$ 
        \ENDFOR
    \end{algorithmic}
\end{algorithm}

\subsection{Traditional Active Learning}
Self-driving companies typically collect large amounts of unlabeled real-world data when operating their vehicles.
Our goal is to select the best subset for labeling to improve model performance.
We assume access to a large, unlabeled pool of examples, $X_{U}$, and an initial subset of labeled examples, $X_{L}^{(0)}$. 
Each example $\V{x} \in X_{U}$ represents an input to our model $f(\V{x})$ and if selected, a labeling oracle returns the ground truth supervision, $\V{y} = L(\V{x})$. 
In the \pnp{} setting, inputs $\V{x}$ represent raw sensor observations and HD Maps, and labels $\V{y}$ represent actor bounding boxes at the current timestep and for the prediction horizon of $T$ seconds.
Each active learning iteration, we select a subset from the remaining unlabeled examples, $Q^{(i)} \subset X_U \setminus X_L^{(i - 1)}$, query the labeling oracle, and add the examples to our labeled set,
\begin{equation}
    X^{(i)}_L = X^{(i - 1)}_{L} \cup Q^{(i)}~~.
\end{equation}
At the end of each iteration, the model can be retrained or fine-tuned with the latest dataset,
\begin{equation}
    \mathcal{D}^{(i)} = \{(\V{x}, \V{y}): \V{x} \in X_{L}^{(i)}\}~~.
\end{equation}

Traditionally, the active learner will select a fixed number of  examples at each iteration, $|Q^{(i)}| = K$.
This implicitly assumes that each example $\V{x} \in  X_{U}$ can be labeled for the same cost, an assumption clearly violated in the \pnp{} setting, which we will relax in the next section. 
While a variety of approaches have been studied for active selection, we focus on methods which produce a scalar score for each example, $S(\V{x}) \in \mathbb{R}$.
Scores represent some notion of informativeness where highly scored examples are believed to be most likely to improve model performance.
For example, measures of model uncertainty, such as entropy, are commonly used (see Section~\ref{model:selection_criteria} for concrete scoring functions for \pnp{}). 
As different models may benefit from different types of examples, most scoring approaches depend on the model's current state.
After scores have been computed for the remaining unlabeled examples, the top $K$ examples can be selected for labeling. 
This process repeats for $N$ active learning iterations and is summarized in Algorithm~\ref{algorithm:traditional_selection}.

\subsection{Fine-Grained Cost-Aware Active Learning}
In this section, we generalize two aspects critical to the \pnp{} setting, allowing for variable labeling costs and fine-grained selection through partial supervision.

\paragraph{Cost-Aware Active Learning:} As a self-driving vehicle operates, the surrounding environment will change, leading to scenes with drastically different labeling costs. Crowded scenes can contain up to hundreds of actors, which are each traditionally labeled with a precise bounding box. 
Sparser scenes, on the other hand, can be labeled with little manual effort. 
To account for these differences, we explicitly model the cost to label each example, $C(\V{x})$. 
At each iteration, rather than select a fixed $K$ examples, the learner is instead given a fixed budget $B$, which cannot be exceeded, 
\begin{equation}
    \sum_{\V{x} \in Q^{(i)}} C(\V{x}) \leq B~~.
\end{equation}

This formulation is a generalization of the previous setting, which  can be recovered by setting $C(\V{x}) = 1$ for all examples and $B = K$.
In practice, labeling cost for \pnp{} examples can be accurately modeled as a linear function of the number of actors in the scene as most annotation time is spent drawing detailed bounding boxes for each actor.
This new formulation requires modifications to our selection algorithm, since high scoring examples $S(\V{x})$ may also have high costs $C(\V{x})$. Therefore, rather than sorting by score, we can select examples with the highest value,
$V(\V{x}) = \frac{S(\V{x})}{C(\V{x})}.$

In practice, the cost of labeling an example $C(\V{x})$ is unknown until it has been labeled. Therefore, for selection, we approximate the cost using the number of detections after NMS as a proxy. Once an example is selected for labeling, its true cost is known, and active selection can continue iteratively until the budget is reached.

Since examples in the \pnp{} setting represent large scenes, scores $S(\V{x})$ and costs $C(\V{x})$ can vary significantly as scenes can have few to many actors, each contributing to the total score and cost. 
As a consequence, coarse-grained scoring will be suboptimal as scenes may contain regions with high score and low cost (e.g., a single car performing a rare U-Turn) and other regions with low score and high cost (e.g., a parking lot filled with many static vehicles). 
This motivates the need for more fine-grained scoring and selection. 
Next, we describe modifications to support partially labeled scenes, which will enable fine-grained selection for better performance in the cost-aware active learning setting.

\paragraph{Partially Labeled Scenes:} 
We generalize the labeling process to allow for partial labeling.
Along with the flexibility it will provide for active selection, this setting is also realistic in practice. 
Even as entire scenes are labeled today, annotation platforms often decompose work into smaller subtasks, which can be more easily distributed and validated across a labeling team. 
As a simple extension, platforms could support querying labels for  only particular regions in the scene. 
To support partial labels, we redefine an example $\V{x}_R$ as the scene augmented with a labeling region $R$,
\begin{equation}
    \V{x}_R = (\V{x}, R)~~.
\end{equation}
Given the set of labels for the entire scene $\V y  = L(\V x)$ and a region $R$, each actor's bounding box label $\V y_{i}$ will either be fully contained in $R$, completely outside of $R$, or partially inside of $R$. 
For simplicity, we assume that if \textbf{any} part of the bounding box $\V y_i$ of an actor is inside $R$ then it will be provided as a label. 
In practice, this translates to labelers annotating all actors, even those that are only partially visible in the queried region of interest. 
More formally, the labeling oracle returns labels for an example $\V{x}_R$, 
\begin{equation}
    L(\V{x}_R) = \{\V y_i : \V y_i \in R \text{ and } \V y_{i} \in L(\V x) \}~~.
\end{equation}

\begin{figure}[t]
    \centering
    \vspace{-0.5cm}
    \includegraphics[width=1.0\linewidth]{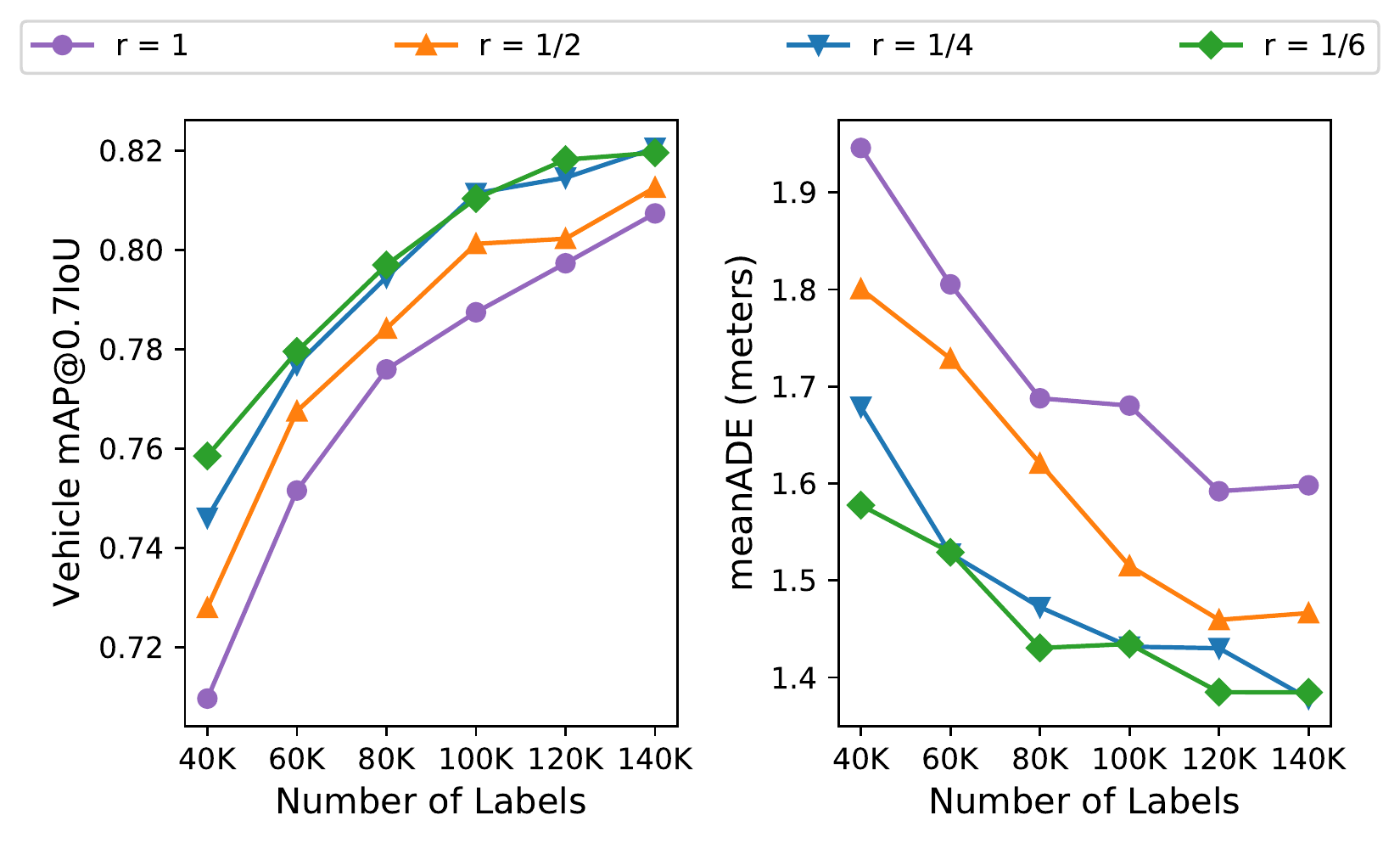}
    \caption{{\bf \pnp{} from Partial Supervision:} \pnp{} performance when trained on partial labels at varying densities, $r$.}
    \vspace{-0.3cm}
    \label{fig:random_regions_pnp}
\end{figure}

\paragraph{Training from Partial Supervision:} We adapt training to support partial supervision by applying the loss only on the labeled region, $R$.
Importantly, we do not alter the network input $\V{x}$, since we do not want to bias the network by changing the input statistics. 
Therefore, the forward pass remains unchanged, $\hat{\V y} = f(\V{x})$. Then, when computing the loss, we only consider the labels that we have received in $R$,
\begin{equation}
    \mathcal{L}(\V y, \hat{\V y} , R) = \ell_{B}(R) + \sum_{\V y_i \in R}\ell_{P}(\V y_i, \hat{\V y}_i)~~.
\end{equation}
Here, $\ell_{P}(\cdot, \cdot)$ represents traditional multi-task perception and prediction losses applied over the positive examples in $R$ and $\ell_{B}(R)$ represents a ``background'' loss which encourages the network not to output detections for negative regions in $R$.
For example, in our experiments, $\ell_{P}$ includes a probabilistic prediction loss,  a bounding box regression loss and cross-entropy on positive examples, whereas $\ell_{B}(R)$ represents the hard negative mining loss, sampling only negative anchors from $R$. Please see the supplementary materials for further details.

\paragraph{Fine-Grained Selection:} 
Without restrictions on $R$, there are infinite regions to consider for a given scene. 
Therefore, in order to efficiently score and select regions for labeling, we consider the set obtained by discretizing the entire scene into a rectangular grid. Specifically, we divide each example $\V x$ into $HW$ non-overlapping regions, 
\begin{equation}
    \V{x}_R = (\V x, R_{h, w}), h = 1 \dots H, w = 1 \dots W~.
\end{equation}
By setting $H = W = 1$, we obtain a single region for each scene and recover the original formulation. 
As $H$ and $W$ increase, candidate regions become smaller, providing the learner more fine-grained precision for selection.
Most steps of the selection process can remain unchanged. 
Scoring functions now operate over examples augmented with regions, $S(\V x_R)$ returning a score that only considers network predictions in $R$. Similarly, only the cost of labeling the queried region $C(\V x_R)$ is incurred when selecting $\V{x}_{R}$.

\begin{figure}[t]
    \centering
    \vspace{-0.5cm}
    \includegraphics[width=1.0\linewidth]{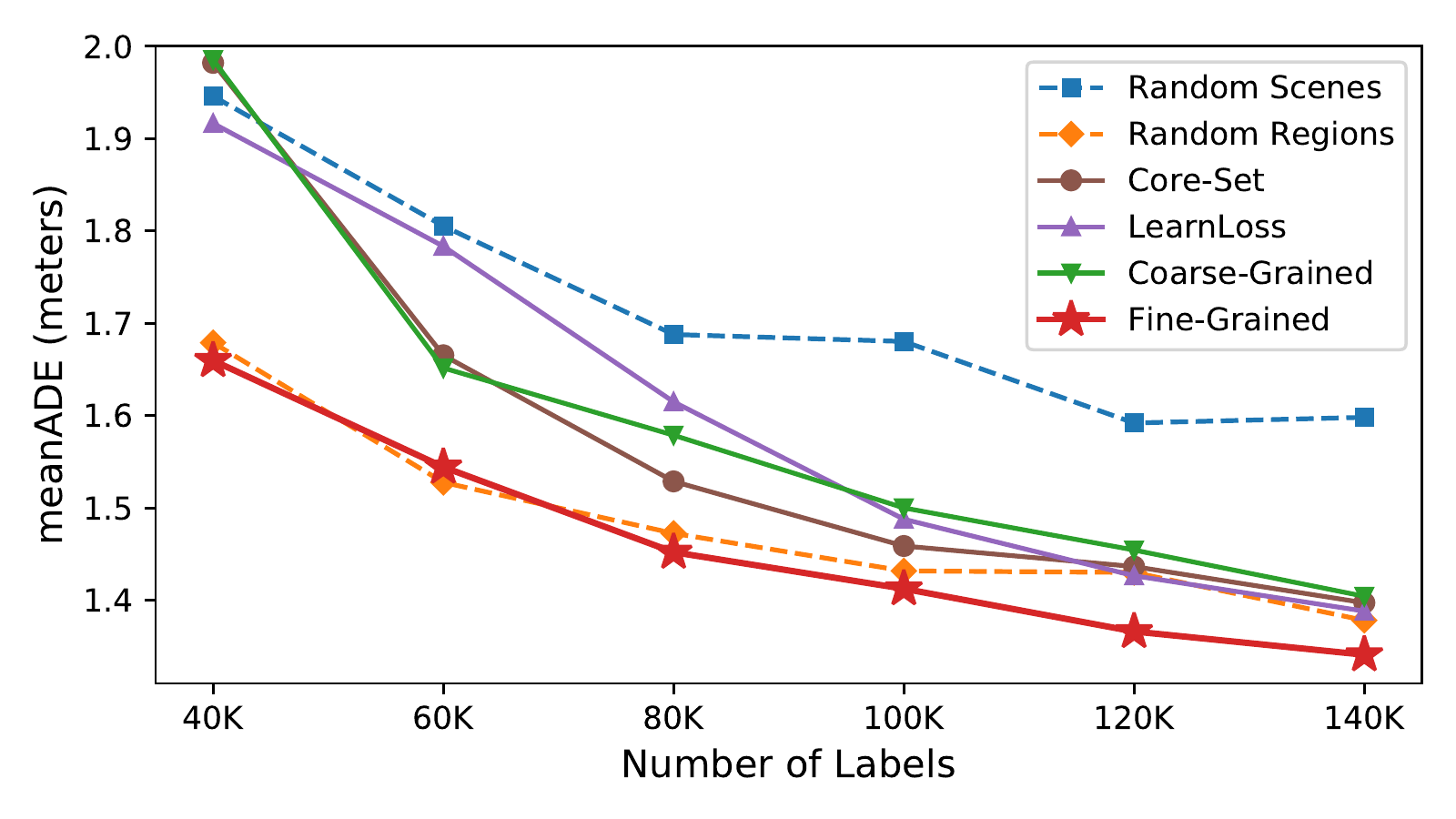}
    \caption{{\bf Active Learning for Prediction:} Performance of various selection approaches over $N = 5$ active learning iterations.}
    \vspace{-0.3cm}
    \label{fig:main_results_ade}
\end{figure}

As regions sizes shrink, we observe that the active learner is more likely to select a large number of scenes, each labeled with very sparse supervision. 
While this dataset would contain many interesting actors, we also find this unconstrained selection results in significantly longer training times as examples are less densely labeled. Additionally, we observe training instabilities due to the imbalances between the amount of supervision available for each example. 
To alleviate these issues, we introduce a sparsity regularizer, which requires that the active learner select a minimum number of positive examples $M$ for any selected scene. 
Therefore, letting $P(\V{x}_R)$ represent the number of positive examples in an example, our new formulation can be summarized by the following optimization problem solved by the learner at each iteration, 
\begin{equation}
    \begin{aligned}
    \max_{Q^{(i)}} \quad & \sum_{\V{x}_R \in Q^{(i)}} S(\V{x}_R) \\
    \textrm{s.t.} \quad & \sum_{\V{x}_R \in Q^{(i)}} C(\V{x}_R) \leq B \\
  \quad & P(\V{x}_R) \geq M ~\forall~\V{x}_R \in Q^{(i)}~~.
    \end{aligned}
\end{equation}

We solve this optimization greedily by first selecting the highest scoring scene remaining, then selecting the highest regions in the scene until at least $M$ actors are labeled. We continue selecting new scenes until the budget is reached.

\begin{table}[t]
\renewcommand{\arraystretch}{0.94}
\vspace{-0.3cm}
\centering
\footnotesize
\begin{tabular}{@{\extracolsep{0.5em}}ccccc}

\\ \toprule
	
\multicolumn{1}{c}{} &
\multicolumn{4}{c}{Prediction (meanADE) $\downarrow$}   \\
	\cmidrule{2-5}
	
	Selection  &
	Straight  &
	Left &
	Right &
	Stationary 

\\\midrule

Random Scenes & 2.89 & 5.31 & 5.68 & 0.22 \\
Random Regions & 2.46 & 4.82 & 4.96 & \textbf{0.20} \\
Core-Set & 2.45 & 4.71 & 5.01 & 0.21 \\
LearnLoss & 2.46 & 4.74 & 4.99 & 0.21 \\
Coarse-Grained & 2.44 & 4.79 & 5.03 & 0.22 \\
Fine-Grained & \textbf{2.29} & \textbf{4.52} & \textbf{4.91} & 0.21 \\

\bottomrule
	
\end{tabular}
\vspace{0.2cm}

\caption{
\textbf{Prediction Performance By High Level Action}
}
\label{table:high_level_actions}
\vspace{-0.3cm}
\end{table}

\subsection{Selection Criteria}
\label{model:selection_criteria}
Our fine-grained, cost-aware active learning formulation introduced above generally supports any approach which provides an informativeness score $S(\V x_R)$ per example.
In this work, we focus on uncertainty-based approaches, an extremely common active learning paradigm based on the assumption 
that training on uncertain examples are most likely to improve future performance. 
As this approach depends on a model's characterization of uncertainty, we first describe our probabilistic \pnp{} model, followed by possible uncertainty-measures that can be used for scoring. 

\paragraph{Model:} Following \cite{Luo2018FastAF}, we jointly train a model for both perception and prediction from LiDAR and HD map inputs.
A model which naturally characterizes uncertainty over predictions is desirable, as these uncertainty estimates provide a useful measure of informativeness for scoring.  
Therefore, we leverage the output representation of \cite{cui2019multimodal}, using a mixture of $K$ Gaussians to represent the distribution of each actor's future positions. 
For simplicity, independence is assumed between timesteps of the prediction horizon, allowing the distribution to be factorized over time.
Thus the likelihood of a particular actor trajectory, $\V y_i$, can be written as,
\begin{equation}
    p(\V y_i) = \sum_{k = 1}^{K} \pi_k \prod_{t = 1}^{T} \mathcal{N}\left(\V y_i ; \mu_k^{t}, \M \Sigma_k^{t} \right)~~,
\end{equation}
where $\mathcal{N}$ is the pdf of a 2D multivariate Gaussian with parameters $\mu_k^{t}$, $\M \Sigma_k^{t}$, and $\pi_k$ represent Gaussian mixture weights.
These parameters, for each detected actor, are predicted by a deep neural network trained with negative log likelihood (see Section~\ref{sec:experiments} for more details). With knowledge of the model, we now introduce measures of uncertainty to use as selection criteria. Due to the multi-task nature of the task, we present separate selection criteria for the detection and prediction task. 
In practice, a mix of both can be used to ensure performance improves across both tasks.

\begin{table}[t]
\setlength{\tabcolsep}{0.03em}
\renewcommand{\arraystretch}{0.94}
\vspace{-0.3cm}
\centering
\footnotesize
\begin{tabular}{@{\extracolsep{0.2em}}ccccccccccc}

\\ \toprule

	Selection  &
	Collision $\downarrow$  &
	L2 Human  $\downarrow$&
	Lat. acc. $\downarrow$ &
	Jerk $\downarrow$&
	Progress $\uparrow$&

\\ 
&
(\% up to 5s) &
($m$ @5s) &
($m/s^2$) &
($m / s^3$) &
($m$ @ 5s) 
	
\\\midrule
	Random Scenes &
	5.02 &
	5.89 &
	2.80 &
	2.67 &
	33.46 &
\\'
Random Regions  &
	5.07 &
	5.71 &
	2.70 &
	2.47 &
	33.65
\\ 
Core-Set &
	5.14 &
	5.72 &
	2.65 &
	2.45 &
	33.63 
\\ 
LearnLoss &
	5.15 &
	5.74 &
	2.68 &
	2.47 &
	33.61
\\ 
Coarse-Grained &
	5.17 &
	5.71 &
	2.67 &
	2.44 &
	\textbf{33.81} 
\\ 
Fine-Grained &
\textbf{4.63} &
\textbf{5.56} &
\textbf{2.62} &
\textbf{2.38} &
33.68
\\ 

\bottomrule
	
\end{tabular}
\vspace{0.1cm}

\caption{
	\textbf{Downstream Planning Performance}
}
\label{table:planning_metrics}
\vspace{-0.3cm}
\end{table}

\paragraph{Detection Entropy:} 
We focus on characterizing the uncertainty over the model's classification predictions for each anchor. 
For classification tasks, the uncertainty is typically estimated by calculating the entropy of the model's predicted probabilities. 
Given anchors $a \in \mathcal{A}$ with associated probabilities $p_{a}$, the entropy of the predictions are given by, 
\begin{equation}
    H_{D}(\mathcal{A}) = -\sum_{a \in \mathcal{A}} p_{a} \log p_{a} + (1 - p_{a}) \log(1 - p_{a})~~.
\end{equation}
To compute a region's score, we assume independence between anchors and sum the entropies of anchors in $R$. 

\begin{figure*}[t]
    \centering
    \vspace{-0.5cm}
    \includegraphics[width=1.0\linewidth]{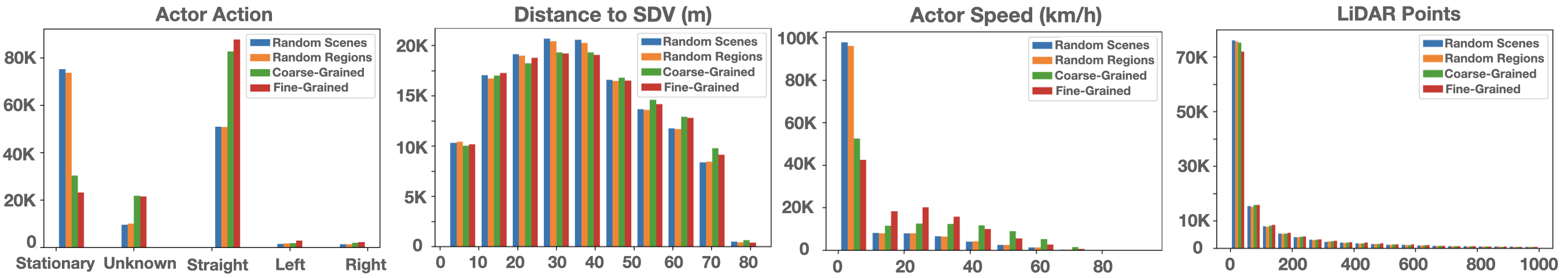}
    \caption{{\bf Selection Statistics:} Statistics of the labels selected by each active selection approach at the final active learning iteration.}
    \vspace{-0.3cm}
    \label{fig:dataset_statistics}
\end{figure*}

\paragraph{Prediction Entropy:} 
Computing prediction entropy naturally depends on the output representation of the model. 
Our model outputs a Gaussian mixture for each predicted actor.
Unfortunately, there is no known closed form solution to computing this distribution's entropy \cite{Huber2008OnEA}. 
Therefore, we are required to estimate the entropy via approximations. 
We explored various approximations, including a sample-based monte-carlo estimate, but all performed similarly to or worse than an approximation via the entropy of the discrete categorical distribution induced by the mixture weights $\pi_k$, 
\begin{equation}
    H_P(\V y_i) = - \sum_{\pi_k} \pi_k \log \pi_k~~.
\end{equation}
Intuitively, this approximation is well-suited to capture cases where the model is uncertain between multiple possible modes, which is likely representative of the true entropy of distribution. 
We use this approximation due to its simplicity and computational efficiency while providing similar performance.
Finally, assuming independence across predicted actors, we sum the entropies of all predictions in a region to obtain the final score for the region.
\section{Experiments}
\label{sec:experiments}
In this section, we analyze the effects of partial labeling and fine-grained active selection.
First, we explore partial labeling independent of active selection and observe significant benefits from sparsely labeled datasets under a fixed labeling budget.
Next, we explore the improvements provided by active selection for prediction. 
We find that a simple prediction entropy combined with fine-grained active selection outperforms various traditional scene-based approaches.
More detailed analysis shows that fine-grained selection enables the learner to better oversample labels exhibiting complex driving behaviors, resulting in better performance on these challenging behaviors in the test-set.
In practice, we are most interested in the effects of these improvements on the downstream motion planning task, where we find significant improvements across most metrics.
Finally, we observe similar improvements for perception when using detection entropy as the selection criterion.
Overall, our experiments demonstrate that partial labeling and fine-grained active selection significantly improve detection, prediction and downstream planning performance.

\begin{figure*}[t]
    \centering
    \vspace{-0.5cm}
    \includegraphics[width=1.0\linewidth]{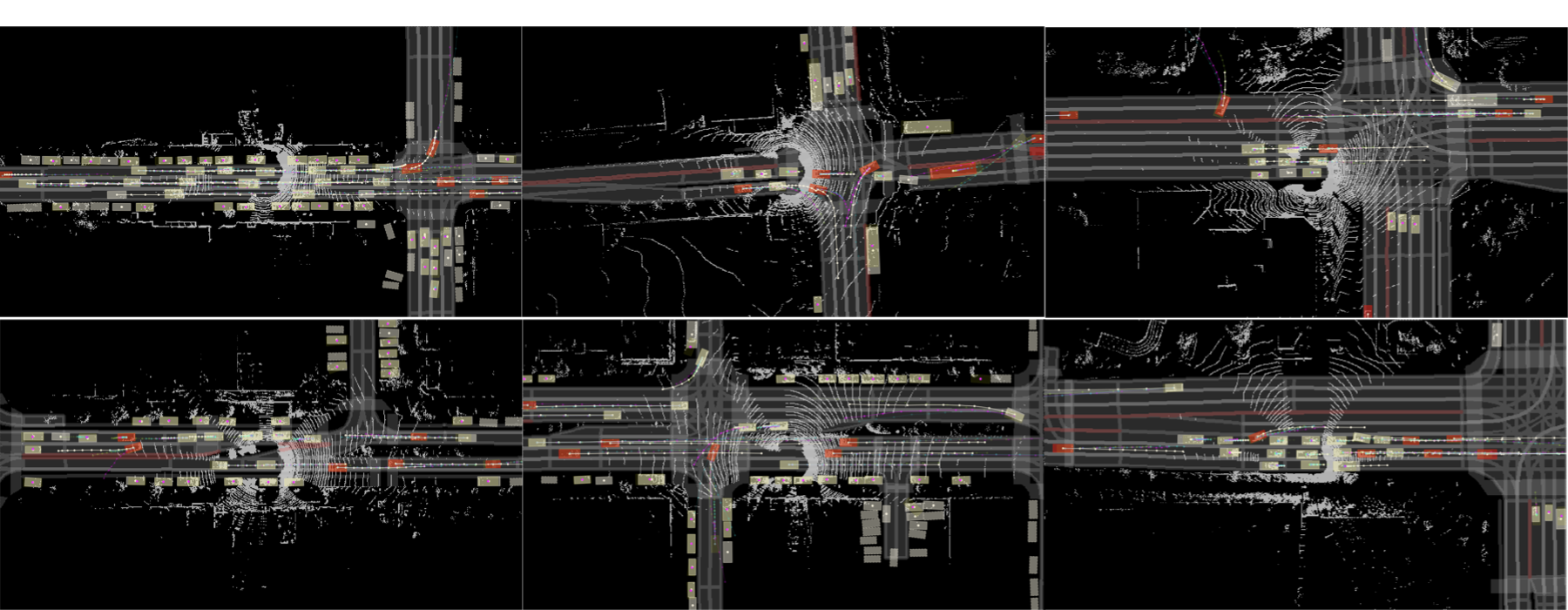}
    \caption{{\bf Qualitative Examples:} Regions selected by fine-grained active selection. We visualize all labels and color those in regions selected by the active learner in \textbf{red}. Selected regions tend to have moving vehicles performing interesting actions (e.g., U-Turns).}
    \vspace{-0.3cm}
    \label{fig:qualitative_examples}
\end{figure*}

\paragraph{Dataset:} For our experiments, we leverage a real-world large-scale self-driving dataset collected across multiple cities in North America.
To simulate the active learning setting, we follow standard practice in active learning research and treat the large labeled dataset as if it were an unlabeled pool.
Then, active learning approaches select from this pool containing 100K scenes, with roughly 2 million actors. 
For each scene, we have access to LiDAR sweeps recorded at 10Hz with a localized HD map given as input to the model.

\paragraph{Metrics:} We evaluate perception and prediction performance 
at each active learning iteration. The model is re-trained on its current set of labels and evaluated on a seperate held-out test set. The test-set is held constant across all approaches and contains traditional fully labeled scenes.
While there exist many metrics for \pnp{}, we focus on mAP@0.7 for detection and meanADE for prediction as they are most commonly employed.
Additionally, we include metrics divided by high level action as well as the performance of the downstream planning task. 

\paragraph{Implementation Details:} For model implementation, we follow the exact details of the Gaussian mixture baseline from \cite{casas2020implicit}, an implementation of MTP \cite{cui2019multimodal} for the joint perception and prediction setting. 
To support partial labeling, we find it is necessary to use \texttt{sum} instead of \texttt{mean} to reduce losses in a batch, ensuring actors in less densely labeled scenes are not up-weighted relative to those in more densely labeled scenes. 
All models are trained for 50 epochs, using budgeted training \cite{Li2020BudgetedTR} for the learning rate schedule. 
For region scoring and selection, we use $H = W = 20$ to discretize the entire scene into $400$ rectangular regions. Finally, when using sparsity regularization, we set $M = 5$.

\subsection{\pnp{} from Partial Supervision}

\paragraph{Experimental Setup:} 
To test the effects of partial labeling, we randomly select labels at varying levels of ground-truth density per scene. 
To ensure a fair comparison, the total labeling budget is fixed across densities.
Specifically, let $r$ be the labeling density. 
When $r = 1$, we recover the traditional fully labeled setting.
When $r = \frac{1}{2}$, we first sample scenes randomly and then sample regions from half of each scene, providing labels only for selected regions.
Notice that since we fix the labeling budget, selecting at lower densities $r$, will result in more scenes for the same budget.

\paragraph{Results:} 
The effect of partial labeling on both detection and prediction performance is shown in Figure~\ref{fig:random_regions_pnp}. 
We find that across all dataset sizes, there is a significant increase in performance with lower density datasets.
Benefits appear to saturate around $r = \frac{1}{4}$, as the sparser labeling density $r = \frac{1}{6}$ does not provide further improvements. 
Performance gains can be explained by the fact that more sparsely labeled datasets naturally include supervision from more scenes, improving the model's ability to generalize.
Our results suggest to optimize \pnp{} performance under a fixed labeling budget, the best strategy is to sparsely label scenes rather than label them in their entirety.
However, we note that in practice there is a tradeoff between labeling at lower densities and model training times, as datasets with less dense supervision must be trained for more iterations. We explore this further in our sparsity regularization experiments.

\subsection{Fine-Grained Active Selection for Prediction}

\paragraph{Experimental Setup:} 
In this experiment, we evaluate active learning approaches to improve prediction performance.
For each method, we sample an initial labeled set $X_L^{(0)}$ of 40K vehicles from $X_U$. 
To ensure that final datasets contain scenes with similar density of supervision, the initial data for fine-grained methods are partially labeled scenes, whereas coarse-grained approaches sample full scenes.
For fair comparison, we fix the labeling budget at each active learning iteration. 
Specifically, for each of the $N = 5$ active learning iterations, the learner is given a budget of $B = 20$K vehicles. 

\paragraph{Additional Baselines:} 
We compare fine-grained active selection to full scenes selected randomly (\texttt{Random Scenes}), partially labeled scenes at density $r = \frac{1}{4}$ selected randomly (\texttt{Random Regions}), full scenes selected by prediction entropy (\texttt{Coarse-Grained}), and two additional active learning baselines adapted to the \pnp{} setting. 
We compare against a recent uncertainty-based approach which learns to predict the loss of unlabeled examples, which we refer to as \texttt{LearnLoss} \cite{Yoo2019LearningLF}.
We train the loss prediction module to predict only the prediction loss (not perception related losses) from the model's intermediate features and re-tune hyperparameters, resulting in a margin $\xi = 1.0$ and loss prediction weight $\lambda = 0.001$.
We additionally compare against the common diversity-based approach of \texttt{Core-Set} selection \cite{Sener2018ActiveLF}. 
Rather than score examples independently, the method seeks to select a representative sample based on distances between examples in a learned space.
Following common practice, we leverage the learned feature representations of the network to compute distances between examples.
For efficiency, we leverage the k-Greedy center variant of the algorithm from the paper, which we find is most commonly used in practice.

\paragraph{Prediction Performance:} Results are shown in Figure~\ref{fig:main_results_ade}. 
All active selection techniques offer significant improvements over random selection. 
Interestingly, despite large differences in the selection criteria (e.g., uncertainty-based vs. diversity-based), scene-based approaches achieve similar performance, indicating that gains may be saturated due to the inflexibility of selecting entire scenes. 
Surprisingly, simply labeling random regions appears to perform better than or similar to many of the coarse-grained active learning approaches. 
Finally, fine-grained selection offers the best performance.
While the improvements may appear to be relatively small, we recall that aggregate prediction metrics are averaged over more than 1M actors in the test-set and may hide large differences between the specific behaviors of the prediction models, calling for more detailed analysis.

\paragraph{Performance By High Level Action:} 
The advantages of fine-grained selection becomes more apparent through more detailed metrics.
In Table~\ref{table:high_level_actions}, we break down the prediction performance by action: driving straight, turning left, turning right, and stationary. 
We notice that differences between selection algorithms become more apparent across actions associated with more difficult predictions (i.e., all non-stationary actions). 
These results are explained by the fact that fine-grained entropy selects more unpredictable moving actors compared to other selection approaches. 

\paragraph{Selection Statistics:} 
Figure~\ref{fig:dataset_statistics} contains histograms of the statistics of the labels selected by each method computed at the final iteration of active learning. \texttt{Core-Set} and \texttt{LearnLoss} are omitted due to similarities with \texttt{Coarse-Grained}. For each approach, we compute the histograms based on label metadata, including their high level action (driving straight, left, right, stationary), the vehicle speed, distance to the SDV, and the number of LiDAR points contained inside the bounding box. 
As expected, we notice that active-selection methods tend to sample more non-stationary vehicles and vehicles further from the SDV. 
This effect is more apparent for fine-grained selection methods due to the additional flexibility provided by the partially labeled setup. 
One potential downside of fine-grained selection is that it will be biased towards regions with actors detected by the current model. 
While this leads to sampling more visible labels (i.e., labels with more LiDAR points), we do not see this affect model performance. 

\begin{figure}[t]
    \centering
    \vspace{-0.7cm}
    \includegraphics[width=1.0\linewidth]{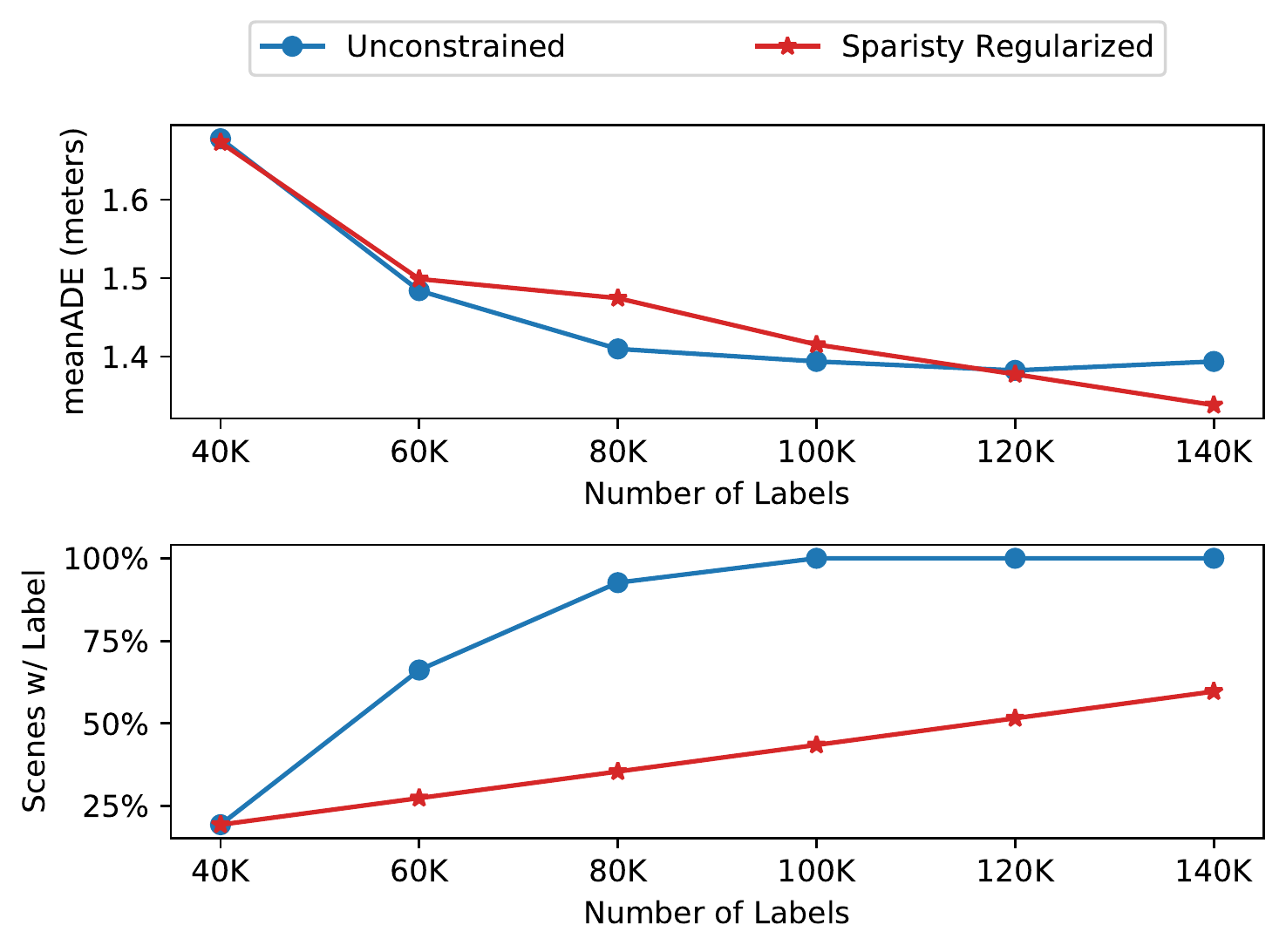}
    \caption{{\bf Sparsity Regularization:} The effect of sparsity regularization on performance (top) and training time (bottom).}
    \label{fig:sparsity_regularization}
    \vspace{-0.3cm}
\end{figure}

\paragraph{Planning Performance:} 
Following \cite{casas2020implicit}, we evaluate downstream performance on motion planning, computing the collision rate, L2 error, lateral acceleration, jerk, and progress of a planner which utilizes the predictions from each trained model, shown in Table~\ref{table:planning_metrics}. 
We notice that fine-grained selection outperforms all baselines across all planning metrics, except progress, for which the differences are not significant.
This demonstrates that, given a fixed budget, fine-grained active selection can improve downstream planning, which is ultimately most important for self-driving.

\paragraph{Qualitative Examples:} Figure~\ref{fig:qualitative_examples} shows qualitative examples of regions selected from fine-grained selection. Selected regions  tend to include vehicle labels (shown in red) with moving actors, actors at intersections, or actors performing odd maneuvers (e.g., U-Turn in the top-right example). Parked and  non-moving actors are rarely selected.

\paragraph{Sparsity Regularization Ablation:} In Figure~\ref{fig:sparsity_regularization}, we ablate the effect of sparsity regularization on performance and number of scenes selected.
At early iterations, we find unconstrained selection outperforms the sparsity regularized approach. 
This is expected, as the unconstrained approach has the freedom to  select a larger set of scenes, each with less supervision. 
However, at later iterations, we observe the unconstrained selection performance degrades relative to sparsity constrained selection. 
We believe this is caused by the imbalances between the supervision available for each scene in the unconstrained selection dataset, as we  have found empirically that these imbalances can lead to degraded performance.
Beyond performance, there is an additional, perhaps more important, benefit of sparsity regularization. Since the active learner must select at least $M$ actors per scene, the number of scenes in the dataset grows linearly with each iteration.
Alternatively, in the unconstrained approach, the dataset size explodes at early iterations until there is at least one label for every scene in $X_U$. 

\begin{figure}[t]
    \centering
    \vspace{-0.5cm}
    \includegraphics[width=1.0\linewidth]{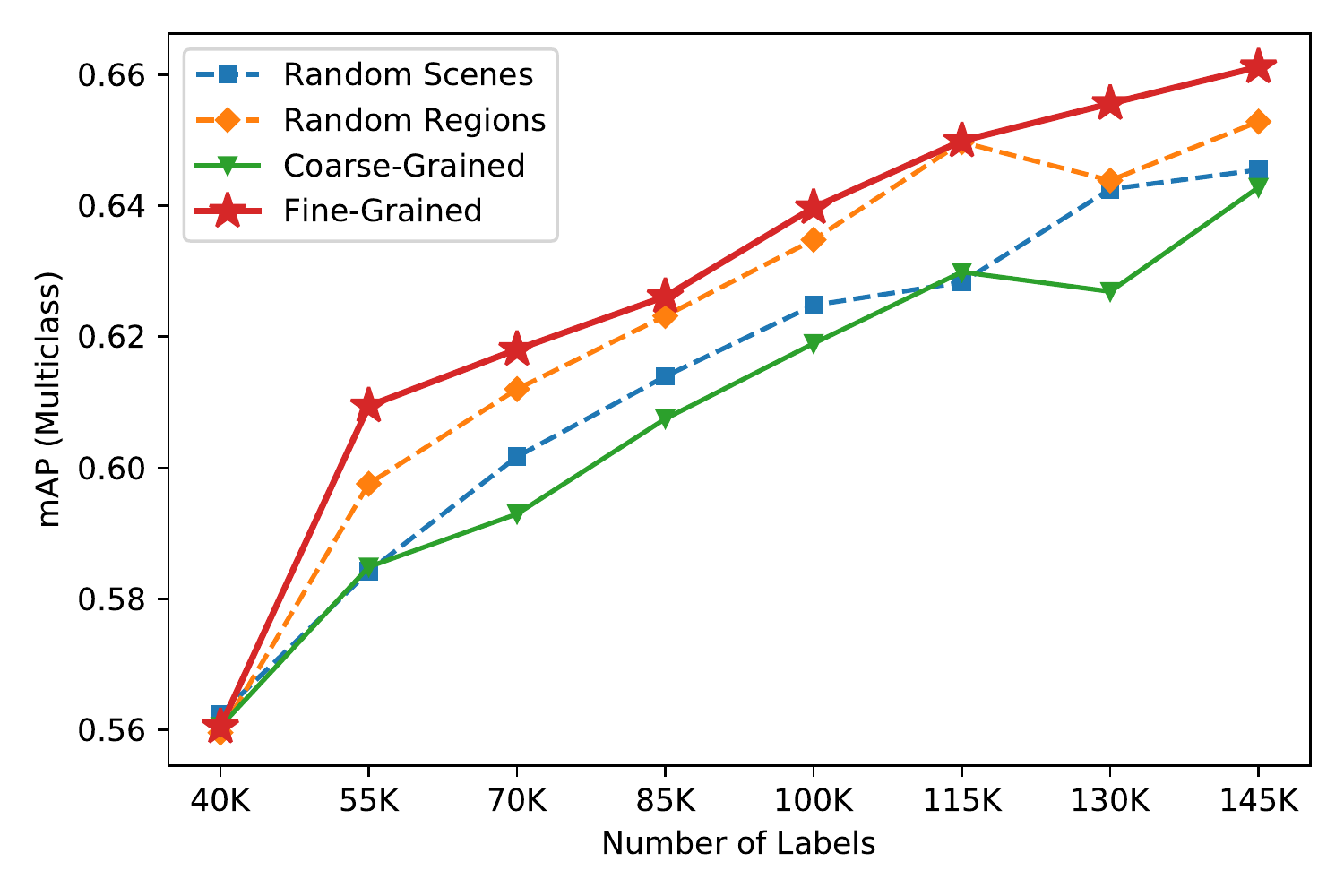}
    \caption{{\bf Active Learning for Perception:} Multi-class detection performance for $N = 7$ iterations of Active Learning.}
    \vspace{-0.3cm}
    \label{fig:active_learning_detection}
\end{figure}

\subsection{Fine-Grained Active Selection for Perception}

\paragraph{Experimental Setup:} 
We additionally experiment with leveraging detection entropy to select examples most likely to improve perception performance.
We follow a similar experimental setup, replacing prediction entropy with detection entropy for selection.
As all methods perform similarly when evaluated only on vehicle detection, we evaluate on the more challenging multi-class setting where cyclists and pedestrians must be detected.

\paragraph{Results:} 
Similar to the prediction setting, results in Figure~\ref{fig:active_learning_detection} show fine-grained selection is most effective.
Interestingly, the performance of coarse-grained selection is similar to random scenes, likely explained by an averaging effect of summing the entropy of predictions over the entire scene.
\section{Conclusion}
In this paper, we studied active learning techniques to intelligently select examples to label from large collections  of unlabeled self-driving data logs for perception and prediction models. 
We found the traditional active learning setting ill-suited and introduced generalizations to account for variable labeling costs and enable fine-grained selection through partially labeled scenes.
In our experiments, we found significant improvements from partial labeling without any active selection, and further gains across perception, prediction and downstream planning by leveraging fine-grained active selection.
Our results demonstrate that the dominant paradigm of labeling entire self-driving scenes may not be most efficient under a fixed budget and that fine-grained selection is likely required for maximal efficiency.
In practice, the best labeling policy should not only optimize performance under a fixed labeling budget, but also account for model training times, 
generalization to new architectures, and robustness to long-tailed events.
We hope our analysis inspires future work on more complex selection criteria designed for these additional considerations. 

{\small
\bibliographystyle{ieee_fullname}
\bibliography{paper}
}

\end{document}